\pdfoutput=1
\documentclass[11pt]{article}
\usepackage[final]{acl}
\usepackage{times}
\usepackage{latexsym}
\usepackage[T2A,T1]{fontenc}
\usepackage{float}
\usepackage[utf8]{inputenc}
\usepackage{CJKutf8}
\newcommand{\textru}[1]{{\fontencoding{T2A}\selectfont #1}}
\usepackage{booktabs}
\usepackage{pifont}
\usepackage{microtype}
\usepackage{inconsolata}
\usepackage{graphicx}
\usepackage{fontawesome5}
\usepackage{amssymb}
\usepackage{enumitem}
\usepackage{hyperref}
\usepackage{natbib}
\usepackage{pgfplots}
\usepackage{pgfplotstable}
\usepackage[table]{xcolor}
\usepackage{multirow}
\usepackage{makecell}
\pgfplotsset{compat=1.18}
\usepackage{amsmath}
\usepackage{cleveref}
\usepackage[most]{tcolorbox}

\title{\textsc{LogSigma} at SemEval-2026 Task 3: Uncertainty-Weighted Multitask Learning for Dimensional Aspect-Based Sentiment Analysis}

\author{Baraa Hikal\textsuperscript{1}, Jonas Becker\textsuperscript{1,2,*}, Bela Gipp\textsuperscript{1,*} \\
  \textsuperscript{1}University of G\"{o}ttingen, Germany; \textsuperscript{2}LKA NRW, Germany, \textsuperscript{*}Advisory Role \\
  {\tt \{baraa.hikal, jonas.becker, gipp\}@uni-goettingen.de}}

\begin{document}
\maketitle

\begin{abstract}
This paper describes \textsc{LogSigma}\footnote{\faGithub\ \href{https://github.com/baraahekal/SemEval2026-LogSigma}{\texttt{Source Code}}}, our system for SemEval-2026 Task~3: Dimensional Aspect-Based Sentiment Analysis (DimABSA). Unlike traditional Aspect-Based Sentiment Analysis (ABSA) that predicts discrete labels, DimABSA requires predicting continuous Valence-Arousal (VA) scores on a 1--9 scale. A key challenge is that Valence and Arousal exhibit different prediction difficulties across languages and domains. We address this through learned homoscedastic uncertainty \citep{8578879}, where the model learns task-specific log-variance parameters ($\log \sigma^2$) that automatically balance each regression objective. Combined with language-specific encoders and multi-seed ensembling, \textsc{LogSigma} achieves 1\textsuperscript{st} place on five datasets across both tracks. The learned variance weights vary substantially across languages due to differing Valence-Arousal difficulty profiles---from $0.66\times$ for German to $2.18\times$ for English---demonstrating that optimal task balancing is language-dependent and cannot be determined \textit{a priori}.
\end{abstract}

\section{Introduction}

 \begin{figure*}[t]
      \centering
      \includegraphics[width=\textwidth]{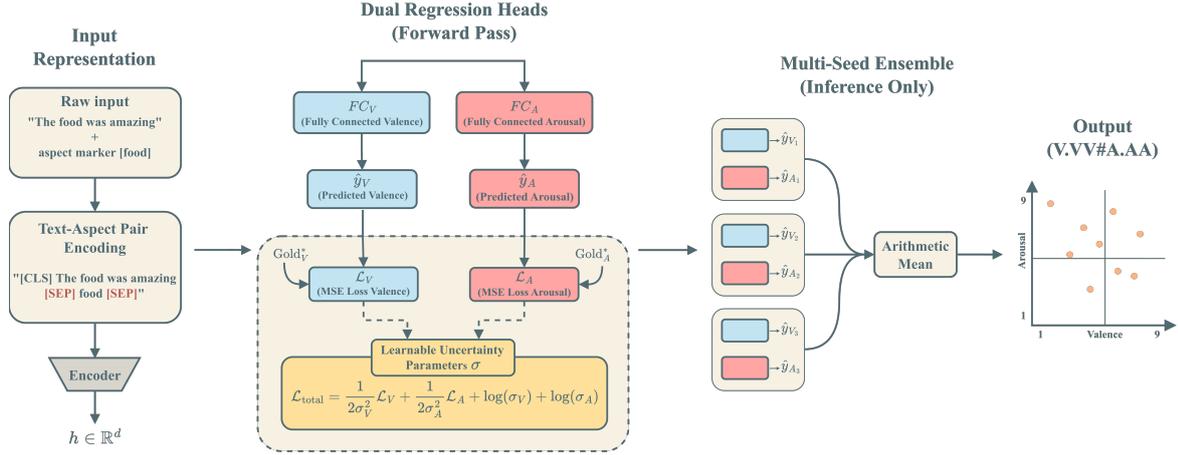}
      \caption{Overview of \textsc{LogSigma}. \textbf{Left:} Input text and aspect are encoded as \texttt{[CLS] text [SEP] aspect [SEP]} and passed through a language-specific pretrained encoder
  to obtain $\mathbf{h} \in \mathbb{R}^d$. \textbf{Middle:} Dual fully-connected heads predict Valence and Arousal independently; during
  training (dashed region), learnable log-variance parameters $s = \log \sigma^2$ balance the MSE losses via precision weighting. \textbf{Right:} At inference, predictions from three seeds are
  averaged to produce the final \texttt{V.VV\#A.AA} output.}
      \label{fig:system}
  \end{figure*}

The SemEval-2026 Task~3 introduces Dimensional Aspect-Based Sentiment Analysis (DimABSA) \citep{yu-etal-2026-semeval}, which reformulates traditional aspect-based sentiment analysis as a
regression problem \citep{lee2026dimabsabuildingmultilingualmultidomain}. Instead of predicting categorical sentiment polarity (positive, negative, neutral) \citep{wilson2009recognizing}, systems must predict continuous
Valence-Arousal (VA) scores on a 1--9 scale \citep{nicolaou2011continuous}, where Valence captures sentiment polarity and Arousal captures emotional intensity \citep{mohammad2016sentiment}. This dimensional
representation preserves fine-grained distinctions lost in categorical classification \citep{reed2016learning}---for instance, ``I love this dish'' and ``the food is acceptable''
both express positive sentiment but differ substantially in intensity \citep{diener1985intensity}. The task spans two tracks: Track~A covers user reviews across six languages \citep{lee2026dimabsabuildingmultilingualmultidomain}
(English, Japanese, Russian, Tatar, Ukrainian, Chinese), while Track~B addresses stance detection across five languages \citep{becker2026dimstancemultilingualdatasetsdimensional} (German, English, Nigerian Pidgin,
Swahili, Chinese).

A key challenge in DimABSA is that Valence and Arousal exhibit different prediction difficulties---Arousal is consistently harder across all languages
(average Pearson correlation gap of 0.29 across all 15 datasets).
Accurate arousal prediction is critical for downstream applications such as crisis detection, brand monitoring, and content moderation, where emotional intensity determines response urgency \citep{mohammad2016sentiment}.
The natural fix---manually tuning per-task loss weights---requires knowing the optimal V/A ratio in advance \citep{chen2018gradnorm}, yet this ratio varies up to $3\times$ across languages (within a single seed) and cannot be estimated without labeled data. Rather than manually tuning per-task loss weights \citep{jha2020adamt}, We propose \textsc{LogSigma}, a system that learns
task-specific log-variance parameters ($\log \sigma^2$) via homoscedastic uncertainty \citep{8578879}, automatically balancing the contribution of
each regression objective. We combine this with language-specific pretrained encoders and multi-seed ensembling.

The learned uncertainty weights encode linguistic structure: related languages converge to near-identical loss landscapes (English and Pidgin within
$\Delta \leq 0.001$ despite per-language training), unrelated languages diverge up to $3\times$ within a single seed, and the directional stability of the learned ratio tracks
dataset-level V/A difficulty separation (Figure~\ref{fig:sigma_comparison}).
\textsc{LogSigma} achieves 1\textsuperscript{st} place on five datasets across both tracks, and language-specific encoders outperform a single multilingual model by 8.2\% on Track~A.

\textbf{Our core contributions are:}
\begin{enumerate}[leftmargin=*, itemsep=2pt, topsep=4pt]
  \item We introduce \textsc{LogSigma}, a multitask learning framework that combines learned homoscedastic uncertainty for automatic V/A loss balancing with language-specific encoders and multi-seed ensembling (\S\ref{sec:system}).
  \item We show that the learned $\sigma^2$ values act as a zero-cost diagnostic, recovering language similarity, encoder pretraining effects, and dataset heterogeneity without explicit linguistic supervision (\S\ref{sec:analysis}).
  \item We achieve 1\textsuperscript{st} place on five datasets across both tracks, with code publicly available. (\S\ref{sec:results}).

\end{enumerate}


\section{Background}

\paragraph{Task Description.}

  The task comprises two tracks. Track~A (DimABSA) includes three subtasks: Subtask~1 (VA regression given text+aspect), Subtask~2 (triplet extraction), and Subtask~3 (quadruplet extraction). Track~B (DimStance) includes Subtask~1 only. We participate in Subtask~1 for both tracks, as it isolates the core continuous VA regression challenge.

  Track~A covers user reviews across four domains and six languages \citep{lee2026dimabsabuildingmultilingualmultidomain} (Chinese, English, Japanese, Russian, Tatar, and Ukrainian). Track~B covers stance detection across five languages \citep{becker2026dimstancemultilingualdatasetsdimensional, mohammad-etal-2016-semeval} (English, German, Chinese, Nigerian Pidgin, and Swahili). Each dataset provides training and development splits; final evaluation uses a held-out blind test set.
  
  The primary evaluation metric is Root Mean Square Error (RMSE, Equation~\ref{eq:rmse}):
  \begin{equation}
  \small
  \text{RMSE} = \sqrt{\frac{1}{N} \sum_{i=1}^{N} \left[ (\hat{V}_i - V_i)^2 + (\hat{A}_i - A_i)^2 \right]}
  \label{eq:rmse}
  \end{equation}
  where $\hat{V}_i$ and $\hat{A}_i$ denote predictions and lower RMSE indicates better performance.
  We participate in English for Track A and all five languages for Track B.

  Dataset statistics are provided in \Cref{tab:track_a,tab:track_b} in \Cref{app:datasets}. Track~A comprises 10 datasets across 6 languages and 4 domains (892--2,507 instances each), while Track~B has 5 datasets in stance detection (876--1,245 instances). Representative examples across languages and sentiment profiles are in \Cref{tab:examples} in \Cref{app:examples}.

  \paragraph{Related Work.}

  Dimensional sentiment analysis builds on Russell's circumplex model \citep{russell1980circumplex}, representing affect along valence (pleasantness) and arousal (activation) \citep{POSNER_RUSSELL_PETERSON_2005}. Resources like Chinese EmoBank \citep{lee2022chinese, buechel-hahn-2017-emobank} provide VA annotations, and the SIGHAN-2024 shared task \citep{lee-etal-2024-overview-sighan} pioneered dimensional ABSA for Chinese.
  Traditional ABSA focuses on discrete polarity labels \citep{9996141}; DimABSA extends this to continuous dimensions.

  For multi-task learning, \citet{8578879} propose learned homoscedastic uncertainty to balance losses when tasks have different noise levels---we adapt this for VA regression. An alternative is GradNorm \citep{pmlr-v80-chen18a}, which dynamically adjusts gradient magnitudes across tasks; we chose homoscedastic uncertainty for its simplicity, requiring only two scalar parameters rather than per-step gradient monitoring. For regression from text, \citet{akhauri2025performancepredictionlargesystems} propose a generative text-to-text paradigm; we found discriminative fine-tuning substantially more effective for VA prediction (Section~\ref{sec:results}).

  Cross-lingual sentiment has benefited from multilingual transformers \citep{conneau-etal-2020-unsupervised, barbieri-etal-2022-xlm}, though language-specific models often outperform on tasks requiring cultural nuance \citep{zhang-etal-2024-sentiment}, consistent with our findings.

\section{System Overview}
\label{sec:system}

  Figure~\ref{fig:system} illustrates the \textsc{LogSigma} architecture, comprising three stages: (1)~input representation via language-specific encoders,
  (2)~dual regression heads with learned uncertainty weighting during training, and (3)~multi-seed ensemble averaging at inference.

  \subsection{Input Representation}

  Given input text and an aspect term, we
  construct a text-aspect pair encoded as \texttt{[CLS] text [SEP] aspect [SEP]}, following \citet{sun-etal-2019-utilizing} (left panel of Figure~\ref{fig:system}). This sequence is passed through a pretrained transformer
  encoder, producing a hidden representation $\mathbf{h} \in \mathbb{R}^d$ from the \texttt{[CLS]} token.

  \subsection{Dual Regression Heads with Learned Uncertainty}
  \label{sec:uncertainty}

  Two fully-connected layers ($\text{FC}_V$ and $\text{FC}_A$) independently project the shared encoder representation to scalar predictions
  (middle panel of Figure~\ref{fig:system}):
  \begin{equation}
  \hat{y}_V = \mathbf{w}_V^\top \mathbf{h} + b_V, \quad \hat{y}_A = \mathbf{w}_A^\top \mathbf{h} + b_A
  \end{equation}

  \paragraph{The Problem with Fixed Weights.}
  A naive approach minimizes $\mathcal{L} = \mathcal{L}_V + \mathcal{L}_A$, implicitly assuming equal task difficulty. Our experiments show this is violated: Arousal exhibits higher noise across all languages (avg.\ Pearson Correlation Coefficient (PCC) gap 0.29; \Cref{tab:va_gap}), and optimal weights are language-dependent, making manual tuning impractical.

  \paragraph{Learned Homoscedastic Uncertainty.}
  Following \citet{8578879}, we model the observation noise for each task as a learnable parameter. For a regression task with Gaussian likelihood,
  the negative log-likelihood is:
  \begin{equation}
  -\log p(y | \hat{y}, \sigma) \propto \frac{1}{2\sigma^2} \| y - \hat{y} \|^2 + \log \sigma
  \end{equation}
  where $\sigma^2$ represents homoscedastic (input-independent) observation variance. Applying this to both tasks yields our training objective (dashed region in
  Figure~\ref{fig:system}):
  \begin{equation}
  \mathcal{L} = \frac{1}{2\sigma_V^2} \mathcal{L}_V + \frac{1}{2\sigma_A^2} \mathcal{L}_A + \log \sigma_V + \log \sigma_A
  \label{eq:uncertainty_loss}
  \end{equation}

  The $1/\sigma^2$ coefficients act as precision-based weights: a task with higher learned variance (more noise) contributes \textit{less} to the total loss,
  preventing noisy gradients from destabilizing training. The $\log \sigma$ regularization terms prevent the trivial solution $\sigma \rightarrow \infty$.

  \paragraph{Practical Implementation.}
  For numerical stability, we parameterize $s = \log \sigma^2$ rather than $\sigma$ directly:
  \begin{equation}
  \mathcal{L} = \frac{1}{2} e^{-s_V} \mathcal{L}_V + \frac{1}{2} e^{-s_A} \mathcal{L}_A + \frac{s_V + s_A}{2}
  \label{eq:practical_loss}
  \end{equation}
  Parameters $s_V$ and $s_A$ are initialized as $s = \log(u)$, $u \sim \text{Uniform}(0.2, 1.0)$, breaking task symmetry from the start. Both are learned jointly with model weights; the system name \textsc{LogSigma} reflects this parameterization.

  \subsection{Language-Specific Encoders}
  \label{sec:encoders}

  Rather than a single multilingual encoder, we select language-specific pretrained models (Table~\ref{tab:encoders}). We prioritize sentiment-pretrained models where available (Twitter-RoBERTa \citep{liu2019robertarobustlyoptimizedbert, loureiro-etal-2022-timelms} for English and Pidgin), use dedicated monolingual models where available (GBERT \citep{chan-etal-2020-Germans} for German), fall back to related-language encoders for low-resource languages (ruBERT \citep{kuratov2019adaptationdeepbidirectionalmultilingual} for Russian and Tatar), and use XLM-RoBERTa-Large \citep{conneau-etal-2020-unsupervised} only for Swahili, which lacks a monolingual model.

  \begin{table}[t]
  \centering
  \small
  \begin{tabular}{ll}
  \toprule
  \textbf{Language} & \textbf{Encoder} \\
  \midrule
  English & \makecell[l]{\texttt{twitter-roberta-large-}\\\texttt{topic-sentiment-latest}}\textsuperscript{$\dagger$} \\
  Japanese & \texttt{bert-base-japanese-v3} \\
  Russian & \texttt{rubert-base-cased} \\
  Tatar & \texttt{rubert-base-cased}\textsuperscript{$\ddagger$} \\
  Ukrainian & \texttt{ukr-roberta-base} \\
  Chinese & \texttt{chinese-roberta-wwm-ext} \\
  German & \texttt{gbert-large} \\
  Pidgin & \makecell[l]{\texttt{twitter-roberta-large-}\\\texttt{topic-sentiment-latest}}\textsuperscript{$\dagger$} \\
  Swahili & \texttt{xlm-roberta-large}\textsuperscript{$\star$} \\
  \bottomrule
  \end{tabular}
  \caption{Language-specific encoders. ($\dagger$) sentiment-pretrained; ($\ddagger$) related-language proxy; ($\star$) multilingual fallback. This strategy outperforms a single multilingual encoder by 8.2\% on Track~A and up to 28.5\% on Track~B (Section~\ref{sec:results}).}
  \label{tab:encoders}
  \end{table}

  \subsection{Multi-Seed Ensemble}

  For each language, we train the same encoder (Table~\ref{tab:encoders}) three times with different random seeds, producing independent predictions $(\hat{y}_{V_i}, \hat{y}_{A_i})$. The final output is the arithmetic mean:
  \begin{equation}
  \hat{y}_V = \frac{1}{3} \sum_{i=1}^{3} \hat{y}_{V_i}, \quad \hat{y}_A = \frac{1}{3} \sum_{i=1}^{3} \hat{y}_{A_i}
  \end{equation}
  formatted as \texttt{V.VV\#A.AA}. This reduces RMSE by 3--5\% over even the best individual seed by smoothing variance from random initialization, a well-established variance-reduction strategy \citep{NIPS2017_9ef2ed4b} (\Cref{tab:ensemble_ablation} in \Cref{app:ensemble}).

\section{Experimental Setup}

  We fine-tune each encoder independently using AdamW \citep{loshchilov2019decoupledweightdecayregularization} with learning rate $2 \times 10^{-5}$ (except English environmental: $1 \times 10^{-5}$, selected via validation), batch size 16 with 4-step gradient accumulation, and early stopping on validation RMSE (patience~3).
  The log-variance parameters $s_V$ and $s_A$ use a separate learning rate of $5 \times 10^{-2}$---approximately $2{,}500\times$ higher than the encoder---enabling rapid adaptation to task-specific noise levels while the encoder preserves pretrained representations.
  We train three models per configuration with seeds $\{21, 99, 42\}$ (Track~A) and $\{42, 12, 73\}$ (Track~B), averaging predictions across seeds.
  All models are trained on a single NVIDIA L4 or L40S GPU (fp16, PyTorch 2.x, Hugging Face Transformers 4.x);
  Full hyperparameters are in \Cref{tab:hyperparams} in \Cref{app:hyperparams}; an epochs ablation (\Cref{tab:epochs_ablation} in \Cref{app:epochs}) informed our choice of early stopping over fixed epoch counts, and a cross-domain transfer study (\Cref{tab:transfer} in \Cref{app:transfer}) motivated language-specific fine-tuning over Track~A initialization.

\section{Results}
\label{sec:results}

\subsection{Main Results}

\textsc{LogSigma} places 1\textsuperscript{st} on five of seven submitted datasets across both tracks, demonstrating that learned uncertainty weighting with language-specific encoders generalizes from development to blind test (Table~\ref{tab:blind_test}).
On Track~A, we submitted predictions for 2 of 10 datasets and ranked 1\textsuperscript{st} on both: English laptop (1\textsuperscript{st}/30 teams) and English restaurant (1\textsuperscript{st}/33 teams).
On Track~B, we ranked 1\textsuperscript{st} on 3 of 5 datasets---German politics (1\textsuperscript{st}/11), English environmental (1\textsuperscript{st}/14), and Swahili politics (1\textsuperscript{st}/11)---and 2\textsuperscript{nd} on Pidgin politics (2\textsuperscript{nd}/11).
Chinese environmental is our weakest result (7\textsuperscript{th}/13), suggesting that our Chinese-RoBERTa encoder, effective for Track~A review domains, may not transfer as well to environmental stance text.

\begin{table}[t]
\centering
\tiny
\setlength{\tabcolsep}{4pt}
\begin{tabular}{llccc}
\toprule
\textbf{Track} & \textbf{Dataset} & \textbf{System} & \textbf{RMSE\,$\downarrow$} & \textbf{Rank} \\
\midrule
\multirow{5}{*}{A} & \multirow{3}{*}{eng\_laptop}
                    & \textsc{LogSigma}  & 1.241 & 1\textsuperscript{st}/33 \\
                    && Kimi-K2 Thinking  & 2.189 & 32\textsuperscript{nd}/33 \\
                    && Qwen-3 14B        & 2.809 & 33\textsuperscript{rd}/33 \\
\cmidrule{2-5}
                   & \multirow{3}{*}{eng\_restaurant}
                    & \textsc{LogSigma}  & 1.104 & 1\textsuperscript{st}/37 \\
                    && Kimi-K2 Thinking  & 2.146 & 36\textsuperscript{th}/37 \\
                    && Qwen-3 14B        & 2.643 & 37\textsuperscript{th}/37 \\
\midrule
\multirow{15}{*}{B} & \multirow{3}{*}{deu\_politics}
                    & \textsc{LogSigma}  & 1.342 & 1\textsuperscript{st}/12 \\
                    && Mistral-3 14B     & 1.591 & 10\textsuperscript{th}/12 \\
                    && mBERT             & 2.329 & 12\textsuperscript{th}/12 \\
\cmidrule{2-5}
                   & \multirow{3}{*}{eng\_environmental}
                    & \textsc{LogSigma}  & 1.473 & 1\textsuperscript{st}/15 \\
                    && Mistral-3 14B     & 1.643 & 7\textsuperscript{th}/15 \\
                    && mBERT             & 2.699 & 15\textsuperscript{th}/15 \\
\cmidrule{2-5}
                   & \multirow{3}{*}{pcm\_politics}
                    & \textsc{LogSigma}  & 1.127 & 2\textsuperscript{nd}/12 \\
                    && Mistral-3 14B     & 1.739 & 9\textsuperscript{th}/12 \\
                    && mBERT             & 3.215 & 12\textsuperscript{th}/12 \\
\cmidrule{2-5}
                   & \multirow{3}{*}{swa\_politics}
                    & \textsc{LogSigma}  & 1.796 & 1\textsuperscript{st}/12 \\
                    && Mistral-3 14B     & 2.299 & 10\textsuperscript{th}/12 \\
                    && mBERT             & 2.784 & 12\textsuperscript{th}/12 \\
\cmidrule{2-5}
                   & \multirow{3}{*}{zho\_environmental}
                    & \textsc{LogSigma}  & 0.646 & 7\textsuperscript{th}/14 \\
                    && Mistral-3 14B     & 0.740 & 12\textsuperscript{th}/14 \\
                    && mBERT             & 1.276 & 14\textsuperscript{th}/14 \\
\bottomrule
\end{tabular}
\caption{Blind test results with baselines from the dataset papers \citep{lee2026dimabsabuildingmultilingualmultidomain,becker2026dimstancemultilingualdatasetsdimensional}. \textsc{LogSigma} outperforms all baselines across every dataset, ranking 1\textsuperscript{st} on five of seven submitted datasets. See \Cref{app:baseline_comparison} for a discussion of the performance gap.}
\label{tab:blind_test}
\end{table}

\subsection{Ablation Study}
\label{sec:ablation}
To isolate the effect of learned uncertainty, we compare the same encoder with and without uncertainty weighting (Table~\ref{tab:uncertainty_ablation}).
Uncertainty improves RMSE on all 5 same-encoder pairs, with relative RMSE reductions of 0.5--11.2\%.
The improvement is largest where V/A difficulty is most imbalanced (Pidgin: $-$11.2\% RMSE); \Cref{fig:pidgin_progression} in \Cref{app:transfer} shows the Pidgin model progression from multilingual baseline to our final system.
Full per-dataset development results are in \Cref{tab:full_track_a,tab:full_track_b} in \Cref{app:dev_results};

\begin{table}[t]
\centering
\footnotesize
\setlength{\tabcolsep}{3.5pt}
\begin{tabular}{llccc}
\toprule
\textbf{Dataset} & \textbf{Encoder} & \textbf{w/o} & \textbf{w/} & \textbf{$\Delta$} \\
\midrule
jpn\_hotel (A)    & XLM-R-L    & 0.672 & \textbf{0.619} & $-$7.9\% \\
jpn\_finance (A)  & XLM-R-L    & 0.673 & \textbf{0.606} & $-$9.9\% \\
deu\_pol (B)      & GBERT-L    & 0.944 & \textbf{0.940} & $-$0.5\% \\
eng\_env (B)      & Tw-RoBERTa & 1.180 & \textbf{1.086} & $-$8.0\% \\
pcm\_pol (B)      & Tw-RoBERTa & 0.702 & \textbf{0.624} & $-$11.2\% \\
\bottomrule
\end{tabular}
\caption{Uncertainty weighting ablation on the dev set (RMSE\,$\downarrow$). Each row compares the same encoder with and without learned uncertainty. The ``w/'' column includes 3-seed ensembling. Gains are largest for Pidgin ($-$11.2\%), where the V/A difficulty imbalance is greatest.}
\label{tab:uncertainty_ablation}
\end{table}

\subsection{Analysis}
\label{sec:analysis}

\paragraph{English--Pidgin sigma convergence.}
The learned $\sigma^2$ values for English and Pidgin converge to within $\Delta \leq 0.001$ across all three seeds (e.g., seed~42: $\sigma^2_V = 0.91$ for both, $\sigma^2_A = 0.42$ for both), despite per-language training.
This quantitatively confirms that the model perceives these two languages as similarly difficult, consistent with Nigerian Pidgin being an English-based creole (\Cref{fig:eng_pcm_convergence} in \Cref{app:sigma} shows all three seeds).

\paragraph{Arousal is universally harder than Valence.}
Across all 15 datasets, Valence PCC (0.84 avg) consistently exceeds Arousal PCC (0.55 avg), yielding an average gap of +0.29.
The gap is largest for Chinese environmental (+0.44) and Russian restaurant (+0.42), and smallest for Chinese restaurant (+0.14); see \Cref{tab:va_gap} in \Cref{app:va_gap} for the full breakdown.
This aligns with the psychological observation that valence (positive vs.\ negative) maps more directly to lexical sentiment, while arousal (intensity) depends on subtler pragmatic cues \citep{russell1980circumplex, warriner2013norms}.

\begin{figure}[t]
    \centering
    \includegraphics[width=\columnwidth]{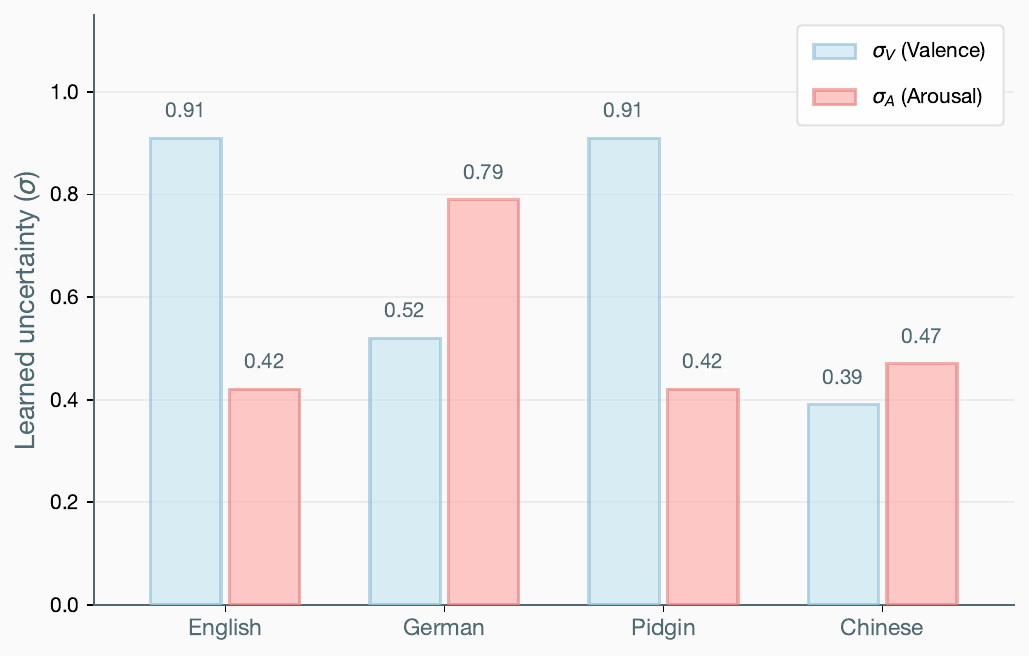}
    \caption{Learned task variance ($\sigma^2$) across Track~B languages (seed~42).
    Higher $\sigma^2$ indicates greater task-specific noise, causing the model to down-weight that objective.
    English and Pidgin show $\sigma^2_V > \sigma^2_A$ (arousal receives higher weight), while German shows the opposite ($\sigma^2_A > \sigma^2_V$).
    English and Pidgin learn near-identical values despite per-language training, confirming their shared V/A difficulty profile.}
    \label{fig:sigma_comparison}
\end{figure}

\paragraph{Learned uncertainty reveals language-specific difficulty.}
Figure~\ref{fig:sigma_comparison} reveals that German is the only language where $\sigma^2_A > \sigma^2_V$ across all seeds, meaning the model down-weights arousal for German political text.
One reason could be the source origin of the German data, which consists of texts by political parties that might publish carefully crafted responses.
This contrasts with other languages, such as Nigerian Pidgin, whose source contains emotionally charged political discussions among voters, yielding higher valence noise ($\sigma^2_V > \sigma^2_A$).
Chinese shows the highest cross-seed instability, with no consistent V/A weighting direction across seeds.
Full per-seed $\sigma^2$ values are in \Cref{app:sigma}; the Track~A extension shows that whether an encoder is sentiment-pretrained or general-purpose---not language family or review domain---is the primary driver of learned V/A difficulty ratios.
Additional experiments including negative results, encoder comparisons, and post-competition ablations are in \Cref{app:encoders}--\Cref{app:error}.

\section{Conclusion}
We presented \textsc{LogSigma}, a system that learns task-specific log-variance parameters to automatically balance Valence and Arousal regression losses for dimensional aspect-based sentiment analysis; combined with language-specific encoders and multi-seed ensembling, it ranks 1\textsuperscript{st} on five of seven submitted datasets across both tracks.
Optimal V/A task weighting varies up to $3\times$ across languages, and sigma profiles converge across related languages (English--Pidgin) and languages sharing the same pretraining regime (Japanese--Russian)---showing that encoder pretraining type is the primary driver of learned V/A difficulty, not language family or review domain.
These findings confirm that adaptive loss balancing discovers language-specific difficulty without manual intervention and doubles as a reproducible dataset-level diagnostic probe.
Future work should explore heteroscedastic uncertainty, $\sigma^2$ transfer to reduce cold-start cost for new languages, and stronger encoders for low-resource settings such as Tatar and Swahili.

\section*{Acknowledgements}
This work was partially supported by the Lower Saxony Ministry of Science and Culture and the VW Foundation.

\bibstyle{acl_natbib}
\bibliography{ref}

\appendix
\crefalias{section}{appendix} 
\onecolumn

\section{Dataset Statistics}
\label{app:datasets}

\begin{table}[H]
\centering
\begin{tabular}{llr}
\toprule
\textbf{Language} & \textbf{Domain} & \textbf{Size} \\
\midrule
\multirow{2}{*}{English} & Restaurant & 2,507 \\
                         & Laptop & 1,869 \\
\midrule
\multirow{2}{*}{Japanese} & Hotel & 1,892 \\
                          & Finance & 1,245 \\
\midrule
Russian & Restaurant & 1,987 \\
Tatar & Restaurant & 892 \\
Ukrainian & Restaurant & 1,124 \\
\midrule
\multirow{3}{*}{Chinese} & Restaurant & 2,156 \\
                         & Laptop & 1,678 \\
                         & Finance & 1,432 \\
\bottomrule
\end{tabular}
\caption{Track~A dataset statistics (6 languages, 4 domains, 10 datasets). Sizes are total aspect-level instances across train and dev splits. Tatar is the smallest dataset (892), contributing to its high prediction difficulty.}
\label{tab:track_a}
\end{table}

Table~\ref{tab:track_b} presents the corresponding statistics for Track~B, which covers stance detection across five languages.

\begin{table}[H]
\centering
\begin{tabular}{llr}
\toprule
\textbf{Language} & \textbf{Domain} & \textbf{Size} \\
\midrule
German & Politics & 1,245 \\
English & Environmental & 987 \\
Nigerian Pidgin & Politics & 1,156 \\
Swahili & Politics & 876 \\
Chinese & Environmental & 1,089 \\
\bottomrule
\end{tabular}
\caption{Track~B dataset statistics (5 languages, stance detection). All datasets are smaller than Track~A, with Swahili being the most resource-constrained (876 instances).}
\label{tab:track_b}
\end{table}

\section{Hyperparameters}
\label{app:hyperparams}

\begin{table}[H]
\centering
\begin{tabular}{lr}
\toprule
\textbf{Hyperparameter} & \textbf{Value} \\
\midrule
Optimizer & AdamW \\
Encoder learning rate & $2 \times 10^{-5}$ \\
Uncertainty lr ($s_V$, $s_A$) & $5 \times 10^{-2}$ \\
Weight decay & 0.01 \\
Batch size / grad.\ accum. & 16 / 4 (eff.\ 64) \\
Max epochs (Track A / B) & 25 / 15 \\
Early stopping patience & 3 \\
Dropout & 0.1 \\
Gradient clipping & 1.0 \\
Warmup & 10\% of total steps \\
Max seq.\ length & 128 \\
Seeds (Track A / B) & \{21, 99, 42\} / \{42, 12, 73\} \\
\bottomrule
\end{tabular}
\caption{Full training hyperparameters. All languages share the same configuration except English environmental (Track~B), which uses a lower encoder learning rate of $1 \times 10^{-5}$. The uncertainty parameters receive a $2{,}500\times$ higher learning rate than the encoder to enable rapid adaptation to task-specific noise levels.}
\label{tab:hyperparams}
\end{table}

\section{Dataset Examples}
\label{app:examples}

\begin{table*}[ht]
\centering
\small
\begin{tabular}{llp{4.5cm}lcc}
\toprule
\textbf{Language} & \textbf{Domain} & \textbf{Text (excerpt)} & \textbf{Aspect} & \textbf{V} & \textbf{A} \\
\midrule
\multicolumn{6}{l}{\textit{\textbf{Track A} --- Multi-Domain Reviews}} \\
\midrule
English  & Restaurant    & \textit{I am so disappointed in the service\ldots}                                                   & service                                                    & 3.38 & 5.75 \\
English  & Laptop        & \textit{I got this laptop and am very impressed\ldots}                                               & laptop                                                     & 8.00 & 8.00 \\
Japanese & Hotel         & \textit{\begin{CJK}{UTF8}{min}客室のトイレが古すぎて臭かったです\end{CJK}\ldots}                    & \begin{CJK}{UTF8}{min}トイレ\end{CJK} \textit{(toilet)}   & 2.62 & 6.50 \\
Japanese & Hotel         & \textit{\begin{CJK}{UTF8}{min}施設のスタッフ\ldots とても丁寧でした\end{CJK}}                       & \begin{CJK}{UTF8}{min}スタッフ\end{CJK} \textit{(staff)}  & 8.00 & 7.83 \\
Russian  & Restaurant    & \textit{\textru{Сервис вполне хороший.}}                                                             & \textru{Сервис} \textit{(service)}                         & 6.50 & 5.00 \\
Chinese  & Restaurant    & \textit{\begin{CJK}{UTF8}{bsmi}真的可以算是\ldots 賣像最差的pizza\end{CJK}}                         & \begin{CJK}{UTF8}{bsmi}賣像\end{CJK} \textit{(appearance)} & 3.33 & 6.67 \\
\midrule
\multicolumn{6}{l}{\textit{\textbf{Track B} --- Stance Detection}} \\
\midrule
German   & Politics      & \textit{Grundversorgung geh\"{o}rt nicht in staatliche Hand.}                                        & Grundversorgung \textit{(basic supply)}                    & 1.90 & 7.40 \\
Pidgin   & Politics      & \textit{A new Nigeria is coming guys.}                                                               & Nigeria                                                    & 7.23 & 4.57 \\
Swahili  & Politics      & \textit{asante kwa taarifa tumezichukua\ldots}                                                       & taarifa \textit{(information)}                             & 7.03 & 5.70 \\
Chinese  & Environmental & \textit{\begin{CJK}{UTF8}{bsmi}人類活動造成嚴重環境污染\end{CJK}\ldots}                             & \begin{CJK}{UTF8}{bsmi}人類活動\end{CJK} \textit{(human activity)} & 3.30 & 6.30 \\
\bottomrule
\end{tabular}
\caption{Representative examples from Track~A and Track~B datasets. Valence (V) and Arousal (A) are gold-annotated scores on a 1--9 scale. The examples illustrate why V/A imbalance is non-trivial: high arousal can co-occur with either high valence (excited approval) or low valence (angry criticism).}
\label{tab:examples}
\end{table*}

\section{Full Development Set Results}
\label{app:dev_results}

\begin{table}[H]
\centering
\begin{tabular}{lcccc}
\toprule
\textbf{Dataset} & \textbf{RMSE\,$\downarrow$} & \textbf{RMSE$_V$} & \textbf{RMSE$_A$} & \textbf{PCC$_V$ / PCC$_A$} \\
\midrule
eng\_restaurant & 0.754 & 0.742 & 0.766 & 0.919 / 0.684 \\
eng\_laptop     & 0.660 & 0.600 & 0.719 & 0.934 / 0.611 \\
jpn\_hotel      & 0.688 & 0.690 & 0.687 & 0.909 / 0.575 \\
jpn\_finance    & 0.650 & 0.801 & 0.500 & 0.828 / 0.489 \\
rus\_restaurant & 1.028 & 0.898 & 1.158 & 0.856 / 0.433 \\
tat\_restaurant & 1.285 & 1.434 & 1.136 & 0.519 / 0.330 \\
ukr\_restaurant & 1.232 & 1.256 & 1.208 & 0.679 / 0.363 \\
zho\_restaurant & 0.532 & 0.582 & 0.482 & 0.848 / 0.711 \\
zho\_laptop     & 0.579 & 0.619 & 0.539 & 0.861 / 0.667 \\
zho\_finance    & 0.439 & 0.464 & 0.414 & 0.814 / 0.498 \\
\midrule
\textbf{Average} & \textbf{0.785} & \textbf{0.809} & \textbf{0.761} & \textbf{0.817 / 0.536} \\
\bottomrule
\end{tabular}
\caption{Track~A full development set results (language-specific encoders + learned uncertainty + 3-seed ensemble). Chinese datasets are consistently easiest (0.44--0.58 RMSE), while low-resource languages are hardest: Tatar (1.29) and Ukrainian (1.23). Arousal RMSE exceeds Valence RMSE on 3 of 10 datasets (both English and Russian); the remaining 7 have RMSE$_V$ > RMSE$_A$, including Japanese finance and all Chinese datasets---see Appendix~\ref{app:va_gap} for analysis.}
\label{tab:full_track_a}
\end{table}

Table~\ref{tab:full_track_b} reports the corresponding Track~B results, where each language uses a different encoder.

\begin{table}[H]
\centering
\begin{tabular}{llccccc}
\toprule
\textbf{Dataset} & \textbf{Encoder} & \textbf{RMSE\,$\downarrow$} & \textbf{RMSE$_V$} & \textbf{RMSE$_A$} & \textbf{PCC$_V$} & \textbf{PCC$_A$} \\
\midrule
deu\_politics      & GBERT-Large      & 0.940 & 0.887 & 0.993 & 0.884 & 0.534 \\
eng\_environmental & Tw-RoBERTa-Large & 1.111 & 1.267 & 0.956 & 0.784 & 0.487 \\
pcm\_politics      & Tw-RoBERTa-Large & 0.624 & 0.567 & 0.681 & 0.977 & 0.746 \\
swa\_politics      & XLM-R-Large      & 1.234 & \multicolumn{2}{c}{\textit{n/a}\textsuperscript{$\dagger$}} & 0.788 & 0.621 \\
zho\_environmental & Zh-RoBERTa       & 0.443 & 0.471 & 0.415 & 0.898 & 0.463 \\
\midrule
\textbf{Average (4 lang)} & & \textbf{0.779} & \textbf{0.798} & \textbf{0.761} & \textbf{0.886} & \textbf{0.557} \\
\bottomrule
\end{tabular}
\caption{Track~B full development set results (best configuration per language). All use learned uncertainty with 3-seed ensembling except Swahili (multilingual ensemble, no uncertainty). The 4-language average excludes Swahili for fair comparison. Pidgin achieves the best RMSE (0.62) despite being a low-resource language, benefiting from the English-pretrained encoder. \textsuperscript{$\dagger$}Per-dimension RMSE not available for Swahili's multilingual ensemble.}
\label{tab:full_track_b}
\end{table}

\section{Learned Uncertainty Parameters}
\label{app:sigma}

\begin{table}[H]
\centering
\begin{tabular}{lcccc}
\toprule
\textbf{Language} & \textbf{Seed} & $\boldsymbol{\sigma^2_V}$ & $\boldsymbol{\sigma^2_A}$ & $\boldsymbol{\sigma^2_V / \sigma^2_A}$ \\
\midrule
\multirow{3}{*}{German}  & 42 & 0.519 & 0.790 & 0.66 \\
                         & 12 & 0.800 & 0.944 & 0.85 \\
                         & 73 & 0.301 & 0.504 & 0.60 \\
\midrule
\multirow{3}{*}{English} & 42 & 0.912 & 0.418 & 2.18 \\
                         & 12 & 0.621 & 0.582 & 1.07 \\
                         & 73 & 0.306 & 0.457 & 0.67 \\
\midrule
\multirow{3}{*}{Pidgin}  & 42 & 0.911 & 0.418 & 2.18 \\
                         & 12 & 0.620 & 0.581 & 1.07 \\
                         & 73 & 0.306 & 0.457 & 0.67 \\
\midrule
\multirow{3}{*}{Chinese} & 42 & 0.394 & 0.471 & 0.84 \\
                         & 12 & 0.888 & 0.228 & 3.90 \\
                         & 73 & 0.769 & 0.495 & 1.55 \\
\bottomrule
\end{tabular}
\caption{Full per-seed learned $\sigma^2$ values (Track~B). Higher $\sigma^2$ indicates greater task-specific noise, causing the model to down-weight that objective. German is the only language where $\sigma^2_A > \sigma^2_V$ across all seeds. English and Pidgin converge to within $\Delta \leq 0.001$ on every seed. Chinese shows the highest cross-seed instability ($\sigma^2_V$: 0.39--0.89), suggesting sensitivity to initialization for this language.}
\label{tab:full_sigma}
\end{table}

\subsection*{Seed Stability Analysis}

We quantify cross-seed stability by the range of $\sigma^2_V / \sigma^2_A$ across three seeds.
A narrow range means the model assigns a consistent relative difficulty to V and~A regardless of initialization.
A wide range means the learned weighting is sensitive to random seed.

\paragraph{German (stable direction).}
The ratio stays between 0.60 and 0.85 across all seeds---always below~1.
Every seed independently concludes that arousal is harder than valence for German political text.
The absolute magnitudes differ ($\sigma^2_V$: 0.30--0.80), but the qualitative weighting direction never changes.
This suggests a structurally consistent difficulty signal in the German data.

\paragraph{English and Pidgin (co-variant instability).}
The ratio ranges from 0.67 to 2.18, and seed~73 crosses the decision boundary (ratio~$< 1$) while seeds~42 and~12 do not.
The weightings are therefore not direction-stable across seeds.
However, English and Pidgin track each other within $\Delta \leq 0.001$ on every seed despite being trained separately.
Whatever initialization-dependent trajectory English takes, Pidgin follows identically---a direct consequence of their linguistic proximity as an English-based creole.

\paragraph{Chinese (high instability).}
Chinese shows the widest range: ratio 0.84--3.90.
Seed~12 assigns $\sigma^2_V = 0.888$ and $\sigma^2_A = 0.228$, strongly down-weighting valence; seed~42 produces a near-balanced assignment (0.394 vs.\ 0.471); seed~73 sits in between (0.769 vs.\ 0.495).
The three seeds therefore optimize under substantially different loss weightings.
This likely reflects the linguistic diversity within the Chinese environmental stance dataset, which may span a wider range of registers than the other four languages.
Even so, Chinese still achieves the best Track~B RMSE (0.44) despite this instability---possibly because the three divergent weighting strategies complement each other in the ensemble.

\begin{figure}[H]
\centering
\includegraphics[width=0.85\textwidth]{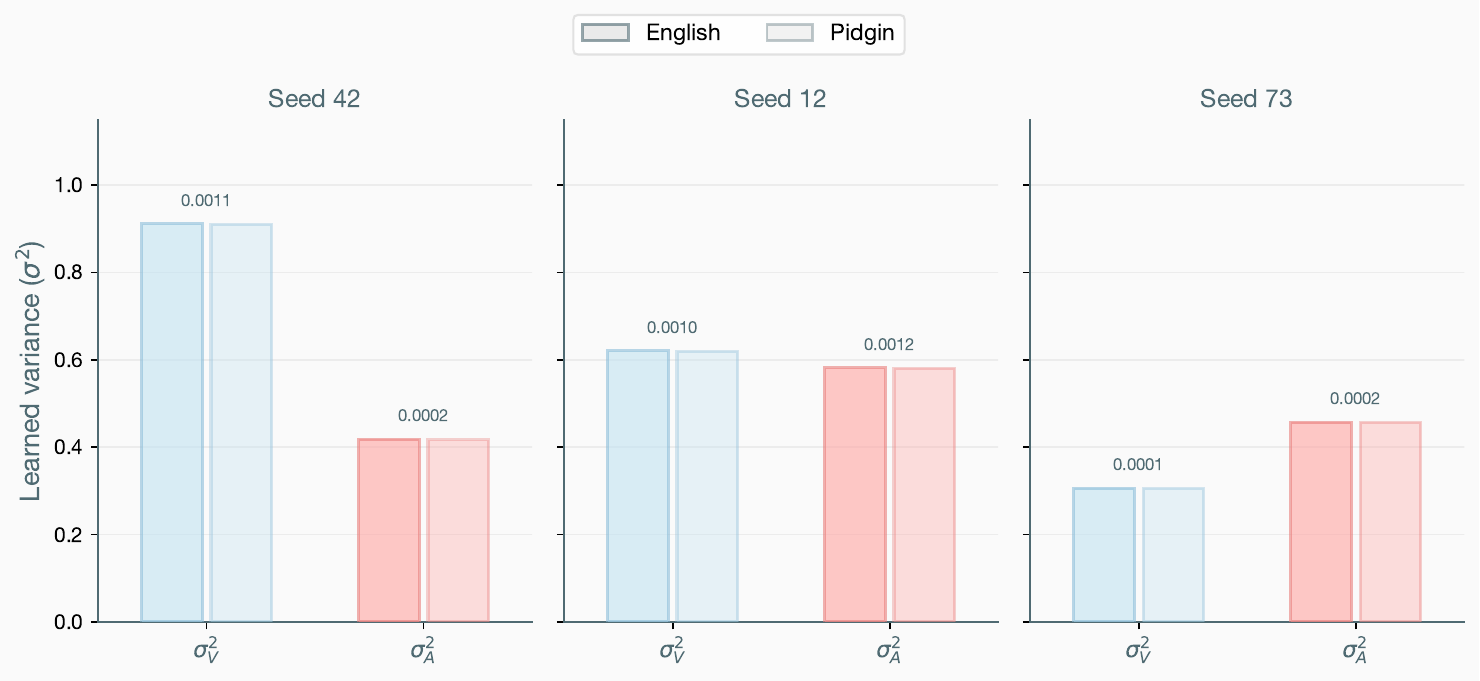}
\caption{English--Pidgin $\sigma^2$ convergence across all three seeds. Despite per-language training on separate datasets, English and Pidgin learn near-identical variance parameters on every seed ($\Delta \leq 0.001$). This quantitatively confirms that the Twitter-RoBERTa encoder perceives these two languages as having identical V/A difficulty profiles, consistent with Nigerian Pidgin being an English-based creole.}
\label{fig:eng_pcm_convergence}
\end{figure}

\subsection*{Track A Sigma Values}

\begin{table}[H]
\centering
\begin{tabular}{lcccc}
\toprule
\textbf{Language} & \textbf{Seed} & $\boldsymbol{\sigma^2_V}$ & $\boldsymbol{\sigma^2_A}$ & $\boldsymbol{\sigma^2_V / \sigma^2_A}$ \\
\midrule
\multirow{3}{*}{English}  & 21 & 0.615 & 0.288 & 2.13 \\
                           & 99 & 0.859 & 0.559 & 1.54 \\
                           & 42 & 0.910 & 0.418 & 2.18 \\
\midrule
\multirow{3}{*}{Japanese} & 21 & 0.354 & 0.752 & 0.47 \\
                           & 99 & 0.443 & 0.525 & 0.84 \\
                           & 42 & 0.392 & 0.469 & 0.84 \\
\midrule
\multirow{3}{*}{Russian}  & 21 & 0.355 & 0.755 & 0.47 \\
                           & 99 & 0.443 & 0.526 & 0.84 \\
                           & 42 & 0.394 & 0.472 & 0.84 \\
\midrule
\multirow{3}{*}{Tatar}    & 21 & 0.355 & 0.755 & 0.47 \\
                           & 99 & 0.443 & 0.527 & 0.84 \\
                           & 42 & 0.394 & 0.472 & 0.84 \\
\bottomrule
\end{tabular}
\caption{Track~A learned $\sigma^2$ values for four languages (seeds 21, 99, 42). English uses restaurant values; Japanese uses finance values (hotel is near-identical). English is the only language with $\sigma^2_V > \sigma^2_A$ across all seeds, and the only Track~A language using a sentiment-pretrained encoder (Twitter-RoBERTa). Japanese and Russian converge to near-identical ratios despite different encoders (BERT-Japanese vs.\ ruBERT), unrelated language families, and different review domains (finance/hotel vs.\ restaurant)---their shared factor is general-purpose encoder pretraining, suggesting that without sentiment pretraining, arousal is consistently the harder target. Russian and Tatar are trivially identical as they share the same ruBERT encoder. Ukrainian and Chinese values were not logged for this run.}
\label{tab:track_a_sigma}
\end{table}

\paragraph{Encoder pretraining type drives divergence.}
Japanese (BERT-Japanese) and Russian (ruBERT) converge to near-identical $\sigma^2_V / \sigma^2_A$ ratios across all three seeds (${\approx}0.47$, $0.84$, $0.84$), despite different encoders, unrelated language families (Japonic vs.\ Slavic), and different review domains (finance/hotel vs.\ restaurant).
Their shared factor is encoder pretraining type: both use general-purpose encoders, and without sentiment pretraining, arousal is consistently the harder target.
English is the sole outlier (ratio ${\approx}1.95$) and the only Track~A language using a sentiment-pretrained encoder; Twitter-RoBERTa's pretraining yields lower valence loss, so the model assigns higher $\sigma^2_V$ and concentrates optimization weight on the harder arousal task.
These findings refine the Track~B English--Pidgin observation: sigma convergence can arise from shared encoder architecture (Russian--Tatar), shared pretraining regime (Japanese--Russian), or shared linguistic and encoder profile (English--Pidgin), and disentangling these factors requires controlled cross-encoder comparisons.

\section{Encoder Comparison (Track~B)}
\label{app:encoders}

\begin{table}[H]
\centering
\begin{tabular}{lccccc}
\toprule
\textbf{Dataset} & \textbf{XLM-R-Large} & \textbf{XLM-R-Base} & \textbf{Afro-XLMR} & \textbf{mDeBERTa} & \textbf{XLM-Politics} \\
\midrule
deu\_politics         & \textbf{1.143} & 1.389 & 1.161 & 1.580 & 1.449 \\
eng\_environmental    & \textbf{1.383} & 1.523 & 1.418 & 1.628 & 1.602 \\
pcm\_politics         & \textbf{0.985} & 1.300 & 0.941 & 1.431 & 1.284 \\
swa\_politics         & 1.464 & \textbf{1.437} & 1.498 & 1.539 & 1.566 \\
zho\_environmental    & \textbf{0.464} & 0.734 & 0.465 & 0.636 & 0.633 \\
\midrule
\textbf{Average}      & \textbf{1.088} & 1.277 & 1.097 & 1.363 & 1.307 \\
\bottomrule
\end{tabular}
\caption{Encoder architecture comparison on Track~B development set (RMSE\,$\downarrow$, all without uncertainty weighting). XLM-RoBERTa-Large wins on 4 of 5 datasets. mDeBERTa-v3-base is the weakest encoder---near-zero arousal PCC on German (0.13) and English (0.18)---suggesting its pretraining does not transfer to VA regression. XLM-Twitter-Politics, despite being pretrained on political sentiment text, underperforms generic XLM-R-Large. XLM-R-Base wins only on Swahili, likely due to less overfitting on this low-resource language.}
\label{tab:encoder_comparison}
\end{table}

\begin{figure}[H]
\centering
\includegraphics[width=0.85\textwidth]{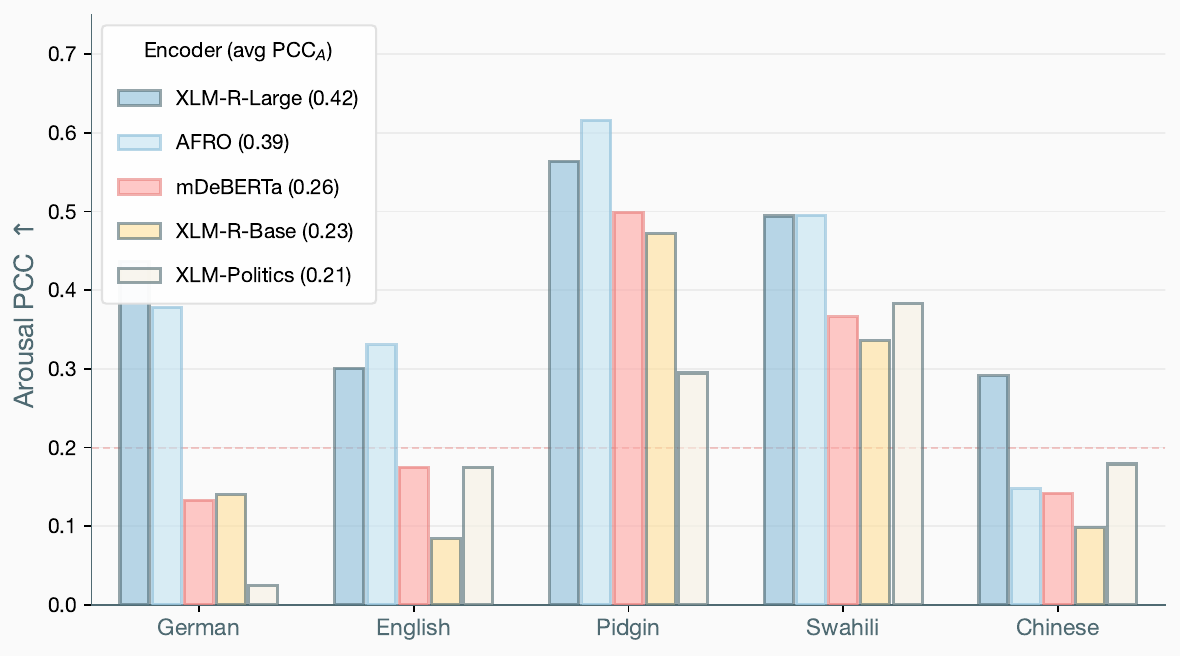}
\caption{Arousal PCC per encoder across Track~B languages. XLM-RoBERTa-Large and Afro-XLMR-Large achieve the highest arousal correlations on most datasets. mDeBERTa and XLM-Twitter-Politics fall below the near-zero threshold (dashed line) on German, demonstrating that their pretraining does not transfer to arousal regression. Pidgin is the only language where all encoders achieve reasonable arousal PCC ($>$0.29), likely because its English-based lexicon benefits all English-pretrained models.}
\label{fig:encoder_pcc_arousal}
\end{figure}

The encoder failures are concentrated on the arousal dimension: mDeBERTa and XLM-Twitter-Politics fall to near-zero arousal PCC on German and English while retaining moderate valence correlation (Figure~\ref{fig:encoder_pcc_arousal}), confirming that pretraining quality matters most for the harder regression target.
The only exception is Swahili, where XLM-R-Base outperforms XLM-R-Large
(1.437 vs.\ 1.464 RMSE)---likely because its lower capacity overfits less on the
smallest dataset (876 instances).

\section{Valence vs.\ Arousal Difficulty Gap}
\label{app:va_gap}

\begin{table}[H]
\centering
\begin{tabular}{lccc}
\toprule
\textbf{Dataset} & \textbf{PCC$_V$} & \textbf{PCC$_A$} & \textbf{Gap ($V - A$)} \\
\midrule
eng\_restaurant    & 0.919 & 0.684 & +0.235 \\
eng\_laptop        & 0.934 & 0.611 & +0.323 \\
jpn\_hotel         & 0.909 & 0.575 & +0.334 \\
jpn\_finance       & 0.828 & 0.489 & +0.339 \\
rus\_restaurant    & 0.856 & 0.433 & +0.423 \\
tat\_restaurant    & 0.519 & 0.330 & +0.189 \\
ukr\_restaurant    & 0.679 & 0.363 & +0.316 \\
zho\_restaurant    & 0.848 & 0.711 & +0.137 \\
zho\_laptop        & 0.861 & 0.667 & +0.194 \\
zho\_finance       & 0.814 & 0.498 & +0.316 \\
\midrule
deu\_politics      & 0.884 & 0.534 & +0.350 \\
eng\_environmental & 0.784 & 0.487 & +0.297 \\
pcm\_politics      & 0.977 & 0.746 & +0.231 \\
swa\_politics      & 0.788 & 0.621 & +0.167 \\
zho\_environmental & 0.898 & 0.463 & +0.435 \\
\midrule
\textbf{Average (15)} & \textbf{0.837} & \textbf{0.546} & \textbf{+0.291} \\
\bottomrule
\end{tabular}
\caption{Valence vs.\ Arousal prediction difficulty across all 15 datasets. Arousal is harder to predict than Valence in every dataset, with an average PCC gap of +0.29. The gap is largest for Chinese environmental (+0.44) and Russian restaurant (+0.42), and smallest for Chinese restaurant (+0.14). Swahili has the smallest Track~B gap (+0.17). However, PCC-based ranking can disagree with RMSE-based ranking: for Japanese finance and all Chinese Track~A datasets, RMSE$_V$ > RMSE$_A$ (Table~\ref{tab:full_track_a}), meaning the model makes larger absolute errors on Valence despite ranking Arousal samples less accurately. This discrepancy arises because PCC measures rank correlation while RMSE measures absolute error.}
\label{tab:va_gap}
\end{table}

\section{Epochs Ablation}
\label{app:epochs}

We compare 5-epoch and 10-epoch training to determine the optimal stopping point per dataset (Table~\ref{tab:epochs_ablation}).

\begin{table}[H]
\centering
\begin{tabular}{lccc}
\toprule
\textbf{Dataset} & \textbf{5 Epochs} & \textbf{10 Epochs} & \textbf{Winner} \\
\midrule
eng\_laptop      & 0.750 & \textbf{0.687} & 10ep \\
eng\_restaurant  & 0.834 & \textbf{0.745} & 10ep \\
jpn\_finance     & \textbf{0.673} & 0.770 & 5ep \\
jpn\_hotel       & \textbf{0.672} & 0.686 & 5ep \\
rus\_restaurant  & 1.002 & \textbf{0.947} & 10ep \\
tat\_restaurant  & \textbf{1.306} & 1.332 & 5ep \\
ukr\_restaurant  & 1.018 & \textbf{0.967} & 10ep \\
zho\_laptop      & \textbf{0.520} & 0.620 & 5ep \\
zho\_restaurant  & \textbf{0.540} & 0.544 & 5ep \\
zho\_finance     & \textbf{0.373} & 0.474 & 5ep \\
\bottomrule
\end{tabular}
\caption{Epochs ablation on Track~A development set (XLM-RoBERTa-Large, RMSE\,$\downarrow$). Five epochs wins on 6 of 10 datasets---all Japanese, all Chinese, and Tatar. Ten epochs wins on English, Russian, and Ukrainian. This suggests that CJK and low-resource languages are more prone to overfitting. Our final system uses early stopping (patience~3) rather than a fixed epoch count to adapt per language.}
\label{tab:epochs_ablation}
\end{table}

\section{Transfer and Adaptation Experiments}
\label{app:transfer}

This section examines three transfer strategies: cross-domain initialization, encoder selection for low-resource languages, and translation-based adaptation.

\paragraph{Cross-domain transfer: Track~A $\to$ Track~B.}
We test whether initializing from a Track~A checkpoint (user reviews) improves Track~B stance performance.
Using XLM-RoBERTa-Large, Track~A initialization improves 4 of 5 languages, with the largest gain on Chinese ($-$12.0\%; Table~\ref{tab:transfer}).
English environmental is the only dataset where transfer hurts (+7.8\%), suggesting that language-specific factors beyond domain mismatch influence transferability.

\begin{table}[H]
\centering
\begin{tabular}{lccc}
\toprule
\textbf{Dataset} & \textbf{Random Init} & \textbf{Track~A Init} & $\boldsymbol{\Delta}$ \\
\midrule
deu\_politics      & 1.143 & \textbf{1.069} & $-$6.5\% \\
eng\_environmental & \textbf{1.383} & 1.490 & +7.8\% \\
pcm\_politics      & 0.985 & \textbf{0.901} & $-$8.5\% \\
swa\_politics      & 1.464 & \textbf{1.362} & $-$6.9\% \\
zho\_environmental & 0.464 & \textbf{0.409} & $-$12.0\% \\
\bottomrule
\end{tabular}
\caption{Cross-domain transfer from Track~A to Track~B, using XLM-RoBERTa-Large (RMSE\,$\downarrow$). Track~A initialization improves 4 of 5 languages (up to $-$12\% on Chinese). English environmental is the only dataset where transfer hurts (+7.8\%), despite Chinese environmental benefiting---suggesting that language-specific factors beyond domain mismatch influence transferability. This experiment uses a single XLM-R model, not our final language-specific setup.}
\label{tab:transfer}
\end{table}

\begin{figure}[H]
\centering
\includegraphics[width=0.85\textwidth]{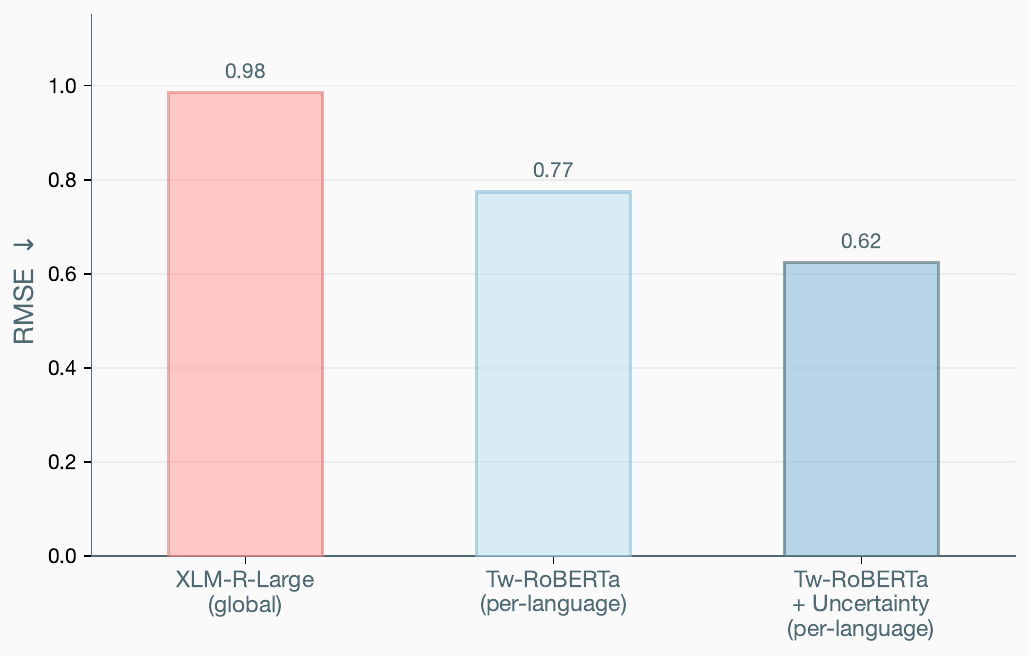}
\caption{Nigerian Pidgin model progression (RMSE on dev set). Switching from multilingual XLM-R (0.98) to English-pretrained Twitter-RoBERTa yields $-$21.4\% RMSE. Adding 3-seed ensembling and learned uncertainty yields a further $-$19.4\% (0.77 $\to$ 0.62), demonstrating that encoder selection, ensembling, and loss design each contribute substantially.}
\label{fig:pidgin_progression}
\end{figure}

\paragraph{Encoder selection for Nigerian Pidgin.}
Nigerian Pidgin (pcm) has no dedicated pretrained model and no Track~A training data, making encoder choice critical.
Figure~\ref{fig:pidgin_progression} traces the full model progression.
Switching from multilingual XLM-R to English-pretrained Twitter-RoBERTa accounts for $-$21.4\% RMSE, while adding 3-seed ensembling and learned uncertainty contributes a further $-$19.4\%.
The uncertainty ablation isolates its contribution at $-$11.2\% (Table~\ref{tab:uncertainty_ablation}), confirming that encoder selection, ensembling, and loss design are complementary.

\begin{figure}[H]
\centering
\includegraphics[width=0.75\textwidth]{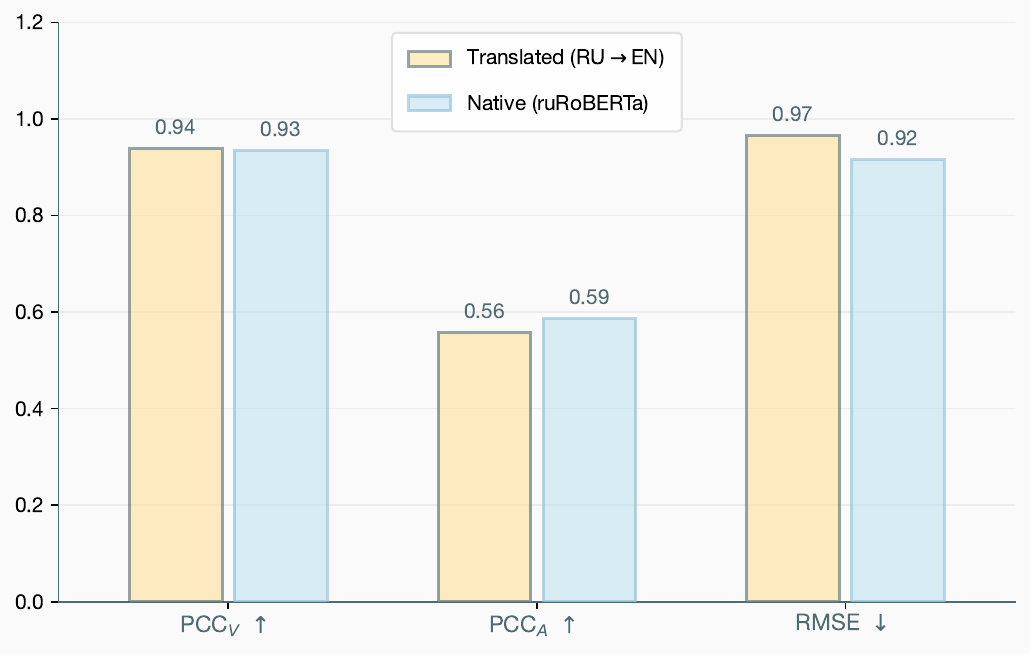}
\caption{Translation vs.\ native model for Russian restaurant. Valence PCC is nearly identical (0.938 vs.\ 0.934), confirming that positive/negative sentiment transfers across languages. However, arousal PCC drops from 0.586 (native) to 0.557 (translated), and overall RMSE increases by 5.4\%. This suggests that arousal is encoded in language-specific features---word choice, emphasis patterns, morphological intensity markers---that translation strips away.}
\label{fig:translation_vs_native}
\end{figure}

\paragraph{Translation.}
Can we bypass language-specific encoders by translating all data to English?
We compare a native Russian model against one trained on machine-translated English data for Russian restaurant.
Valence PCC is nearly preserved (0.938 translated vs.\ 0.934 native), but arousal PCC drops from 0.586 to 0.557 (Figure~\ref{fig:translation_vs_native}).
This confirms that arousal is encoded in language-specific features---word choice, emphasis patterns, morphological intensity---that translation strips away, justifying our language-specific encoder design.

\section{What Did Not Work}
\label{app:negative}

\paragraph{Generative regression.}
To test whether framing VA prediction as sequence generation improves over discriminative models, we evaluate RegressLM \citep{akhauri2025performancepredictionlargesystems}.
We compare a pretrained T5-Gemma backbone (60M params, lr\,=\,2e-5, early stopped at epoch 16) against a from-scratch 2-encoder-2-decoder variant (31.2M params, lr\,=\,5e-5, early stopped at epoch 46); both are trained jointly on all languages with MSE loss.

The pretrained T5-Gemma variant achieves 1.48 avg RMSE, nearly double the 0.78 of our discriminative \textsc{LogSigma} (Table~\ref{tab:regresslm}).
Augmenting its inputs with synthetic opinion terms narrows this gap to 1.34 (\Cref{tab:opinion_ablation} in \Cref{app:post_competition}), yet remains 71\% behind.
The from-scratch variant performs substantially worse: on Japanese, it produces negative PCC, meaning predictions are anti-correlated with gold labels (Table~\ref{tab:from_scratch}).
Pretrained initialization alone accounts for a 15.7\% RMSE reduction (1.59\,$\to$\,1.34), confirming the importance of transfer learning for low-resource multilingual regression.

\begin{table}[H]
\centering

\begin{tabular}{lccc}
\toprule
\textbf{Dataset} & \textbf{RegressLM} & \textbf{\textsc{LogSigma}} & $\boldsymbol{\Delta}$ \\
\midrule
eng\_laptop      & 1.438 & \textbf{0.660} & $-$54.1\% \\
eng\_restaurant  & 1.334 & \textbf{0.754} & $-$43.5\% \\
jpn\_hotel       & 1.333 & \textbf{0.688} & $-$48.4\% \\
jpn\_finance     & 1.523 & \textbf{0.650} & $-$57.3\% \\
rus\_restaurant  & 1.873 & \textbf{1.028} & $-$45.1\% \\
tat\_restaurant  & 2.137 & \textbf{1.285} & $-$39.9\% \\
ukr\_restaurant  & 1.736 & \textbf{1.232} & $-$29.0\% \\
zho\_restaurant  & 1.109 & \textbf{0.532} & $-$52.0\% \\
zho\_laptop      & 1.372 & \textbf{0.579} & $-$57.8\% \\
zho\_finance     & 0.922 & \textbf{0.439} & $-$52.4\% \\
\midrule
\textbf{Average} & \textbf{1.478} & \textbf{0.785} & $-$46.9\% \\
\bottomrule
\end{tabular}
\caption{RegressLM (T5-Gemma, pretrained, 60M params) on Track~A dev set (RMSE\,$\downarrow$), text + aspect input only. $\Delta$ shows the relative RMSE reduction of \textsc{LogSigma} over RegressLM. The opinion-augmented variant is in \Cref{tab:opinion_ablation} in \Cref{app:post_competition}.}
\label{tab:regresslm}
\end{table}

The from-scratch variant (Table~\ref{tab:from_scratch}) performs substantially worse, producing negative PCC on Japanese datasets.

\begin{table}[H]
\centering
\begin{tabular}{lrrr}
\toprule
\textbf{Dataset} & \textbf{PCC$_V$} & \textbf{PCC$_A$} & \textbf{RMSE} \\
\midrule
eng\_laptop      & 0.69  & 0.32  & 1.560 \\
eng\_restaurant  & 0.78  & 0.56  & 1.294 \\
jpn\_finance     & 0.03  & $-$0.03 & 1.610 \\
jpn\_hotel       & $-$0.05 & $-$0.11 & 1.969 \\
rus\_restaurant  & 0.32  & 0.23  & 1.997 \\
tat\_restaurant  & 0.41  & 0.24  & 1.951 \\
ukr\_restaurant  & 0.48  & 0.08  & 1.908 \\
zho\_laptop      & 0.07  & 0.16  & 1.395 \\
zho\_restaurant  & 0.12  & 0.16  & 1.355 \\
zho\_finance     & 0.14  & 0.10  & 0.829 \\
\midrule
\textbf{Average} & \textbf{0.30} & \textbf{0.17} & \textbf{1.587} \\
\bottomrule
\end{tabular}
\caption{From-scratch RegressLM (2E2D, $d_{\text{model}}$ = 512, 31.2M params, lr = 5e-5, early stopped at epoch 46) on Track~A dev set. Japanese datasets produce \textit{negative} PCC on both dimensions, meaning predictions are anti-correlated with gold labels. Even the best language (English restaurant, PCC$_V$ = 0.78) falls far short of pretrained encoders (PCC$_V$ = 0.93 with \textsc{LogSigma}).}
\label{tab:from_scratch}
\end{table}

\paragraph{Alternative encoders.}
mDeBERTa-v3-base \citep{he2023debertav} produces near-zero arousal PCC on German (0.13) and English (0.18), and averages 1.36 RMSE---worst among all encoders tested (\Cref{tab:encoder_comparison} in \Cref{app:encoders}, \Cref{fig:encoder_pcc_arousal}).
XLM-Twitter-Politics, despite being pretrained on political sentiment text, underperforms generic XLM-RoBERTa-Large.
Domain-specific sentiment pretraining does not transfer to VA regression.

\paragraph{Translation.}
Translating Russian to English degrades arousal while preserving valence; see \Cref{app:transfer} for the full analysis.

\section{Post-Competition Experiments}
\label{app:post_competition}

\subsection{Synthetic Opinion Generation}

\paragraph{Motivation: opinion terms drive VA variance.}
Subtask~1 provides text and aspect terms, while opinion annotations are available for all domains except the finance domain in two languages.
An exploratory analysis of the Track~A training data shows that, within the same aspect, opinion terms are the primary source of VA variability (\Cref{tab:va_consistency}).
Grouping instances by aspect alone yields large within-group variance.
Adding opinion to the grouping key substantially reduces this variance: on average, $\sigma$ decreases by \textbf{79.1\%} for Valence and \textbf{55.1\%} for Arousal.
An analysis of variance (ANOVA) further confirms this effect: opinion accounts for nearly all residual within-aspect variance, with $\eta^2_V{=}0.98$ and $\eta^2_A{=}0.95$.
This suggests that VA scores are largely determined by opinion rather than aspect identity.

\paragraph{Illustrative example.}
As a concrete example, the aspect \textit{``food''} in \texttt{eng\_restaurant} spans V\,=\,1.17--8.75 depending on the expressed opinion, where positive opinions such as \textit{``delicious''} yield high scores (8.00) and negative opinions such as \textit{``terrible''} produce low scores (1.17).
Two datasets (jpn\_finance, zho\_finance) originally lack opinion annotations; for these we use synthetic DeepSeek-generated opinions (\S\ref{app:post_competition}).
Synthetic opinions reduce jpn\_finance variance by 85.8\%, on par with ground-truth datasets, while zho\_finance shows a smaller 47.4\% reduction---consistent with the domain shift observed in its RMSE ablation.

\begin{table}[H]
\centering

\begin{tabular}{lcccccc}
\toprule
\textbf{Dataset} & \multicolumn{2}{c}{\textbf{Aspect-only $\sigma$}} & \multicolumn{2}{c}{\textbf{Aspect+Opinion $\sigma$}} & \multicolumn{2}{c}{\textbf{$\Delta$\%}} \\
\cmidrule(lr){2-3}\cmidrule(lr){4-5}\cmidrule(lr){6-7}
 & V & A & V & A & V & A \\
\midrule
eng\_laptop      & 1.28 & 0.85 & 0.33 & 0.41 & $-$74.5 & $-$51.6 \\
eng\_restaurant  & 1.30 & 0.87 & 0.30 & 0.37 & $-$76.8 & $-$57.0 \\
jpn\_hotel       & 0.91 & 0.49 & 0.22 & 0.24 & $-$76.0 & $-$50.9 \\
\rowcolor{gray!15} jpn\_finance$^\dagger$     & 0.92 & 0.43 & 0.13 & 0.12 & $-$85.8 & $-$71.5 \\
rus\_restaurant  & 1.34 & 0.80 & 0.16 & 0.30 & $-$88.2 & $-$61.8 \\
tat\_restaurant  & 1.24 & 0.79 & 0.19 & 0.35 & $-$84.5 & $-$56.2 \\
ukr\_restaurant  & 1.28 & 0.78 & 0.18 & 0.33 & $-$86.1 & $-$57.7 \\
zho\_restaurant  & 0.77 & 0.55 & 0.23 & 0.29 & $-$70.8 & $-$47.4 \\
zho\_laptop      & 0.82 & 0.55 & 0.21 & 0.27 & $-$74.5 & $-$50.3 \\
\rowcolor{gray!15} zho\_finance$^\dagger$     & 0.39 & 0.31 & 0.21 & 0.19 & $-$47.4 & $-$37.9 \\
\midrule
\textbf{Average} & \textbf{1.03} & \textbf{0.64} & \textbf{0.21} & \textbf{0.29} & $\mathbf{-79.1}$ & $\mathbf{-55.1}$ \\
\bottomrule
\end{tabular}
\caption{Within-group standard deviation of VA scores when instances are grouped by aspect alone versus aspect+opinion.
Adding opinion reduces $\sigma_V$ by 47--88\% and $\sigma_A$ by 38--72\%, confirming that opinion is the dominant source of within-aspect VA variance ($\eta^2_V{=}0.98$, $\eta^2_A{=}0.95$).
($\dagger$)~Synthetic DeepSeek-generated opinions; all other datasets use ground-truth labels.}
\label{tab:va_consistency}
\end{table}

\paragraph{Method: DeepSeek few-shot pipeline.}
For the two domains without opinion annotations (jpn\_finance, zho\_finance), we generate synthetic opinion terms via few-shot prompting with DeepSeek-V3.2 \citep{deepseekai2025deepseekv32pushingfrontieropen} (\texttt{deepseek-chat}, temperature 0.1, max 50 tokens).
All other datasets use ground-truth opinion labels from the Track~A training data.

For each text--aspect pair, the model extracts the exact opinion word or phrase, or outputs \texttt{NULL} if none is expressed.
Since the finance domains lack native examples, few-shot prompts draw from related training sets (\texttt{jpn\_hotel} and \texttt{zho\_restaurant}, respectively), balanced across positive, negative, and null categories.
Both languages use native-language system prompts; the English template is shown in Figure~\ref{fig:opinion_prompt}.
\begin{figure}[H]
\centering
\begin{tcolorbox}[
  enhanced,
  colback=white,
  colframe=black!60,
  boxrule=0.5pt,
  arc=2mm,
  left=5pt, right=5pt, top=4pt, bottom=4pt,
  fontupper=\small,
  width=0.95\columnwidth
]
\textbf{\faRobot~System}\\[2pt]
{\small You are an expert at Aspect-Based Sentiment Analysis (ABSA). Given a text and an aspect term, extract the exact opinion word or phrase from the text that expresses sentiment about that aspect. The opinion should be a word or short phrase directly from the text expressing positive, negative, or neutral sentiment. If no explicit opinion is expressed, output ``NULL''. Output ONLY the opinion word/phrase.}

\tcblower

\textbf{\faUser~User}\\[2pt]
{\small%
\textit{Example 1 (positive):} Text: \colorbox{gray!15}{\texttt{\{pos\_text\}}}\\
\hspace*{1em}Aspect: \colorbox{gray!15}{\texttt{\{aspect\}}} $\to$ Opinion: \colorbox{gray!15}{\texttt{\{pos\_opinion\}}}\\[3pt]
\textit{Example 2 (negative):} Text: \colorbox{gray!15}{\texttt{\{neg\_text\}}}\\
\hspace*{1em}Aspect: \colorbox{gray!15}{\texttt{\{aspect\}}} $\to$ Opinion: \colorbox{gray!15}{\texttt{\{neg\_opinion\}}}\\[3pt]
\textit{Example 3 (null):} Text: \colorbox{gray!15}{\texttt{\{null\_text\}}}\\
\hspace*{1em}Aspect: \colorbox{gray!15}{\texttt{\{aspect\}}} $\to$ Opinion: \texttt{NULL}\\[5pt]
Now extract the opinion for:\\
Text: \colorbox{gray!15}{\texttt{\{text\}}}~~Aspect: \colorbox{gray!15}{\texttt{\{aspect\}}}\\
Opinion:}
\end{tcolorbox}
\caption{English system prompt and few-shot template for synthetic opinion generation with DeepSeek. Each of the 6 languages uses an equivalent native-language prompt. Examples cover positive, negative, and null opinion categories.}
\label{fig:opinion_prompt}
\end{figure}

\paragraph{Results: opinion ablation.}
We ablate the impact of including opinion terms—using ground-truth labels for eight datasets and synthetic DeepSeek-generated labels for the two finance domains—using a T5-Gemma baseline (Table~\ref{tab:opinion_ablation}). Appending these opinions to the input reduces average RMSE by \textbf{9.2\%} (1.48 $\to$ 1.34). 

Improvements are consistent across both ground-truth and synthetic subsets, with the largest gains observed in \texttt{zho\_laptop} ($-25.0\%$) and \texttt{rus\_restaurant} ($-17.5\%$). Synthetic opinions also reduced error in \texttt{jpn\_finance} ($-9.5\%$), validating the few-shot extraction strategy for filling annotation gaps. 

Performance did not improve in two cases: \texttt{tat\_restaurant} and \texttt{zho\_finance}. For Tatar, where ground-truth opinions were used, the +3.2\% increase suggests that the baseline model struggled to map native-language opinion terms to VA scores. For \texttt{zho\_finance}, the +6.5\% regression likely stems from domain shift, as few-shot examples were drawn from restaurant data. Although the generative model remains 42\% behind \textsc{LogSigma}, the consistent improvements confirm that opinion terms encode meaningful VA signal, motivating future joint modeling.

\begin{table}[H]
\centering
\begin{tabular}{lccc}
\toprule
\textbf{Dataset} & \textbf{$-$Opinion} & \textbf{+Opinion} & $\boldsymbol{\Delta}$ \\
\midrule
eng\_laptop      & 1.438 & \textbf{1.279} & $-$11.1\% \\
eng\_restaurant  & 1.334 & \textbf{1.188} & $-$10.9\% \\
jpn\_hotel       & 1.333 & \textbf{1.224} & $-$8.2\%  \\
jpn\_finance$^\ddagger$  & 1.523 & \textbf{1.379} & $-$9.5\%  \\
rus\_restaurant  & 1.873 & \textbf{1.545} & $-$17.5\% \\
tat\_restaurant  & \textbf{2.137} & 2.205 & +3.2\%    \\
ukr\_restaurant  & 1.736 & \textbf{1.722} & $-$0.8\%  \\
zho\_restaurant  & 1.109 & \textbf{0.859} & $-$22.5\% \\
zho\_laptop      & 1.372 & \textbf{1.029} & $-$25.0\% \\
zho\_finance$^\ddagger$  & \textbf{0.922} & 0.982 & +6.5\%    \\
\midrule
\textbf{Average} & 1.478 & \textbf{1.341} & $-$9.2\%  \\
\bottomrule
\end{tabular}
\caption{Opinion ablation with RegressLM (T5-Gemma, pretrained, 60M params) on Track~A dev set (RMSE\,$\downarrow$). \textbf{$-$Opinion}: text + aspect only (same baseline as Table~\ref{tab:regresslm}); \textbf{+Opinion}: input includes opinion terms (Ground-Truth for 8 sets; Synthetic for 2 sets marked with $\ddagger$). Inclusion of opinion terms reduces average RMSE by 9.2\%; see Table~\ref{tab:regresslm} for \textsc{LogSigma} comparison.}
\label{tab:opinion_ablation}
\end{table}
\subsection{LLM Fine-tuning}

We evaluate whether scaling to a 4B-parameter LLM improves over our encoder-based system.
To probe the upper performance bounds of Track~A, we fine-tuned Qwen3-4B with two regression heads (5 epochs, MSE loss). As shown in Table~\ref{tab:qwen}, the LLM reduces RMSE by \textbf{21.4\%} on \texttt{rus\_restaurant} compared to \textsc{LogSigma} (0.808 vs.\ 1.028). This suggests that for high-variance domains, the larger parameter count can better separate overlapping VA distributions.

Despite these gains, we did not scale the LLM approach across all 15 datasets due to three key factors. First, we observed diminishing returns: in \texttt{jpn\_finance}, the improvement over \textsc{LogSigma} was negligible (0.645 vs. 0.650), indicating that LLM scale does not always translate to better performance in specialized or lower-variance domains. Second, the fundamental V/A asymmetry remained unresolved; the model still struggled significantly more with Arousal than Valence (1.08 vs. 0.54 in Russian), suggesting this is a data-level challenge rather than a capacity issue. Finally, the extreme computational cost of fine-tuning 4B-parameter models made them impractical for the competition's multilingual requirements, where our encoder-based discriminative models offered a much higher efficiency-to-performance ratio.

\begin{table}[H]
\centering
\begin{tabular}{lrrrc}
\toprule
\textbf{Dataset} & \textbf{RMSE$_V$} & \textbf{RMSE$_A$} & \textbf{RMSE} & \textbf{\textsc{LogSigma}} \\
\midrule
rus\_restaurant & 0.538 & 1.077 & \textbf{0.808} & 1.028 \\
jpn\_finance    & 0.835 & 0.455 & \textbf{0.645} & 0.650 \\
\bottomrule
\end{tabular}
\caption{Post-competition Qwen3-4B fine-tuning (2 regression heads, 5 epochs, MSE loss) on two Track~A dev datasets (RMSE\,$\downarrow$). While Qwen3-4B delivers a substantial improvement in the Russian domain, it exhibits diminishing returns in Japanese finance, leading to our decision to prioritize more efficient encoder models.}
\label{tab:qwen}
\end{table}
\vspace{10em}
\section{Ensemble Ablation}
\label{app:ensemble}

Table~\ref{tab:ensemble_ablation} compares individual seed runs against the 3-seed ensemble for two representative datasets.

\begin{table}[H]
\centering
\begin{tabular}{llccccc}
\toprule
\textbf{Dataset} & \textbf{Track} & \textbf{Seed 1} & \textbf{Seed 2} & \textbf{Seed 3} & \textbf{Ensemble} & $\boldsymbol{\Delta}$ \\
\midrule
eng\_restaurant & A & 1.277 & 1.211 & 1.214 & \textbf{1.153} & $-$4.8\% \\
eng\_environmental & B & 1.218 & 1.201 & 1.251 & \textbf{1.161} & $-$3.3\% \\
\bottomrule
\end{tabular}
\caption{Single-seed vs.\ 3-seed ensemble (RMSE\,$\downarrow$). $\Delta$ is the relative improvement from the best individual seed to the ensemble. Averaging predictions across three seeds consistently improves over even the best single run, smoothing variance from random initialization.}
\label{tab:ensemble_ablation}
\end{table}

\section{Language Difficulty Ranking}
\label{app:difficulty}

Table~\ref{tab:difficulty} ranks all 15 dataset--language combinations by best development RMSE.

\begin{table}[H]
\centering
\begin{tabular}{clllc}
\toprule
\textbf{\#} & \textbf{Language} & \textbf{Track} & \textbf{Domain} & \textbf{Best RMSE\,$\downarrow$} \\
\midrule
1  & Chinese   & A & Finance       & 0.373 \\
2  & Chinese   & B & Environmental & 0.409 \\
3  & Chinese   & A & Laptop        & 0.520 \\
4  & Chinese   & A & Restaurant    & 0.532 \\
5  & Japanese  & A & Finance       & 0.606 \\
6  & Japanese  & A & Hotel         & 0.619 \\
7  & Pidgin    & B & Politics      & 0.624 \\
8  & English   & A & Laptop        & 0.660 \\
9  & English   & A & Restaurant    & 0.686 \\
10 & Russian   & A & Restaurant    & 0.808 \\
11 & German    & B & Politics      & 0.940 \\
12 & Ukrainian & A & Restaurant    & 0.967 \\
13 & English   & B & Environmental & 1.086 \\
14 & Swahili   & B & Politics      & 1.234 \\
15 & Tatar     & A & Restaurant    & 1.285 \\
\bottomrule
\end{tabular}
\caption{All 15 dataset-language combinations ranked by best development RMSE across all configurations tested. Chinese is the easiest language across all domains and tracks (0.37--0.53). Tatar and Swahili are the hardest ($>$1.23), both lacking dedicated pretrained encoders. English environmental ranks 13\textsuperscript{th}---harder than German politics---likely because environmental stance text is distributionally different from the sentiment-rich data that Twitter-RoBERTa was pretrained on. Note: some entries use earlier configurations (e.g., XLM-R-Large) that outperformed the final language-specific system on individual datasets; final system results are in Tables~\ref{tab:full_track_a} and~\ref{tab:full_track_b}.}
\label{tab:difficulty}
\end{table}



\section{Error Analysis}
\label{app:error}

\paragraph{Domain transfer failure.}
Chinese-RoBERTa achieves 0.44--0.58 RMSE on Track~A Chinese reviews (Table~\ref{tab:full_track_a}) but is our weakest submission on Track~B Chinese environmental stance (7\textsuperscript{th}/13 teams, 0.646 blind test RMSE).
Same encoder, same language, different domain: the encoder's sentiment representations do not transfer to stance-detection text.
This tells us that domain-specific pretraining matters as much as language-specific pretraining.

\paragraph{Low-resource encoder gap.}
Tatar (1.285 RMSE) and Swahili (1.234) post our two worst results (Table~\ref{tab:difficulty}).
Both lack dedicated monolingual encoders: Swahili falls back to XLM-R-Large while Tatar uses ruBERT as a Russian proxy.
The gap to the next-hardest dataset (English environmental: 1.086) is large, confirming that the encoder gap---not dataset size alone---is the primary driver.
Developing monolingual encoders for these languages remains the clearest path to improvement.

\section{Comparison Against Dataset Paper Baselines}
\label{app:baseline_comparison}

Table~\ref{tab:blind_test} includes official baselines from the Track~A \citep{lee2026dimabsabuildingmultilingualmultidomain} and Track~B \citep{becker2026dimstancemultilingualdatasetsdimensional} dataset papers: Kimi-K2 Thinking and Qwen-3 14B on Track~A, and Mistral-3 14B and mBERT on Track~B. Three factors explain the performance gap.

\textbf{Track~A baselines use token-based generation.} Both Track~A baselines predict VA scores auto-regressively as text tokens. Kimi-K2 Thinking uses zero/one-shot prompting; Qwen-3 14B is fine-tuned via supervised learning but remains a generative decoder. The Track~A dataset paper explicitly notes that this produces ``a random grid-like VA distribution, likely due to its token-based generation.'' VA regression requires a continuous output space, which auto-regressive generation cannot provide. This structural mismatch---not model capacity---explains why Kimi-K2 Thinking scores 2.189 RMSE (32\textsuperscript{nd}/33) on eng\_laptop and 2.146 (36\textsuperscript{th}/37) on eng\_restaurant, while \textsc{LogSigma} achieves 1.241 and 1.104 on the same sets using direct MSE regression.

\textbf{All baselines apply one model across all languages.} On Track~B, Mistral-3 14B uses a regression head \citep{becker2026dimstancemultilingualdatasetsdimensional}---the same discriminative setup as \textsc{LogSigma}---yet ranks 7\textsuperscript{th}--10\textsuperscript{th} because it uses a single LLM for all five languages. mBERT, a generic multilingual encoder, finishes last on every dataset (e.g., 2.329 vs.\ our 1.342 on German; 3.215 vs.\ our 1.127 on Pidgin). The Track~B dataset paper's own PLM experiments confirm this: XLM-R-Large, the strongest single multilingual encoder baseline, averages 1.641 RMSE across languages---more than twice our 0.779 (4-language average). \textsc{LogSigma} uses dedicated pretrained encoders per language (Table~\ref{tab:encoders}), which alone accounts for 8.2\% average RMSE improvement over the multilingual baseline (\S\ref{sec:ablation}).

\textbf{No baseline addresses V/A difficulty or variance.} Arousal is harder than Valence in every dataset (avg.\ PCC gap of 0.29; Table~\ref{tab:va_gap}). All baselines---whether generative or regression-based---apply uniform loss weighting, implicitly over-weighting the easier Valence objective. \textsc{LogSigma}'s learned parameters ($s_V$, $s_A$) correct this automatically, with the largest gains where the V/A gap is widest: Pidgin ($-$11.2\%), English environmental ($-$8.0\%), Japanese finance ($-$9.9\%) (Table~\ref{tab:uncertainty_ablation}). A further 3--5\% comes from 3-seed ensembling, which no baseline employs (Table~\ref{tab:ensemble_ablation}).

\end{document}